\documentclass[11pt,a4paper]{article}
\usepackage[hyperref]{acl2021}
\usepackage{times}
\usepackage{latexsym}

\usepackage{microtype}

\usepackage{bm}
\usepackage{amsmath}
\usepackage{url}
\usepackage{graphicx}
\usepackage{float} 
\usepackage{subfigure}
\usepackage{multirow}
\usepackage{makecell}
\usepackage{xpinyin}
\usepackage{enumerate}

\usepackage{xcolor}
\usepackage{soul}
\newcolumntype{L}[1]{>{\raggedright\let\newline\\\arraybackslash\hspace{0pt}}m{#1}}
\setul{2pt}{2pt}
\usepackage{booktabs}
\usepackage{amsmath,amsfonts}
\usepackage[utf8]{inputenc}
\usepackage{cleveref}

\crefname{section}{§}{§§}
\Crefname{section}{§}{§§}
\aclfinalcopy 

\usepackage{enumitem}
\setenumerate[1]{itemsep=0pt,partopsep=0pt,parsep=\parskip,topsep=5pt}
\setitemize[1]{itemsep=0pt,partopsep=0pt,parsep=\parskip,topsep=5pt}
\setdescription{itemsep=0pt,partopsep=0pt,parsep=\parskip,topsep=5pt}
\usepackage{arydshln}

\title{Modeling Role-Specific Consistency and Multi-Perspective Dialogue Coherence for Neural Chat Translation}
\title{Modeling Role-Specific Consistency and Language-Wise Dialogue Coherence for Neural Chat Translation}
\title{Modeling Bilingual Conversational Characteristics\\ for Neural Chat Translation}
\author{
  Yunlong Liang\textsuperscript{1}\thanks{ \ \ Work was done when Yunlong Liang was interning at Pattern Recognition Center, WeChat AI, Tencent Inc, China.}  , 
  Fandong Meng\textsuperscript{2}, 
  \textbf{Yufeng Chen}\textsuperscript{1}, \textbf{Jinan Xu}\textsuperscript{1}\thanks{ \ \ Jinan Xu is the corresponding author.}
  \ and \textbf{Jie Zhou}\textsuperscript{2}\\
  \textsuperscript{1}Beijing Key Lab of Traffic Data Analysis and Mining, \\Beijing Jiaotong University, Beijing, China \\
  \textsuperscript{2}Pattern Recognition Center, WeChat AI, Tencent Inc, China \\
  \texttt{\{yunlongliang,chenyf,jaxu\}@bjtu.edu.cn} \\
  \texttt{\{fandongmeng,withtomzhou\}@tencent.com} \\
}
\date{}

\begin{document}
\maketitle
\begin{abstract}

Neural chat translation aims to translate bilingual conversational text, which has a broad application in international exchanges and cooperation.
Despite the impressive performance of sentence-level and context-aware Neural Machine Translation (NMT), there still remain challenges to translate bilingual conversational text due to its inherent characteristics such as role preference, dialogue coherence, and translation consistency. 
In this paper, we aim to promote the translation quality of conversational text by modeling the above properties. 
Specifically, we design three latent variational modules to learn the distributions of bilingual conversational characteristics. Through sampling from these learned distributions, the latent variables, tailored for role preference, dialogue coherence, and translation consistency, are incorporated into the NMT model for better translation. We evaluate our approach on the benchmark dataset BConTrasT (English$\Leftrightarrow$German) and a self-collected bilingual dialogue corpus, named BMELD (English$\Leftrightarrow$Chinese). Extensive experiments show that our approach notably boosts the performance over strong baselines by a large margin and significantly surpasses some state-of-the-art context-aware NMT models in terms of BLEU and TER. 
Additionally, we make the BMELD dataset publicly available for the research community.\footnote{Code and data are publicly available at: \url{https://github.com/XL2248/CPCC}}
\end{abstract}

\section{Introduction}
A conversation may involve participants that speak in different languages (\emph{e.g.}, one speaking in English and another in Chinese). \autoref{fig:case} shows an example, where the English role $R_1$ and the Chinese role $R_2$ are talking about the ``\emph{boat}''. The goal of chat translation is to translate bilingual conversational text, \emph{i.e.}, converting one participant’s language (\emph{e.g.}, English) to another’s (\emph{e.g.}, Chinese) and vice versa~\cite{farajian-etal-2020-findings}. It enables multiple speakers to communicate with each other in their native languages, which has a wide application in industry-level services. 
\textbf{\begin{figure}[t]
    \centering
    \includegraphics[width=0.48\textwidth]{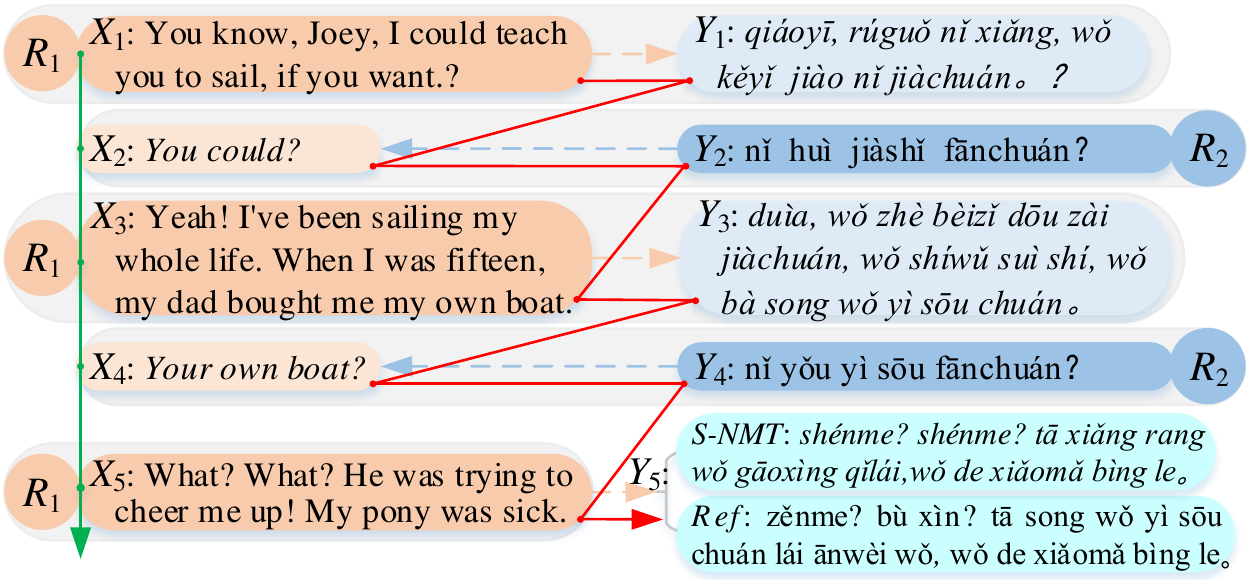}
    \caption{An ongoing bilingual conversation example (English$\Leftrightarrow$Chinese), where the Chinese utterances are presented in pinyin style. $R_i$: Role $i$. The dashed arrows mark the translation direction. The green and red arrows represent the monolingual and bilingual conversation flow, respectively. Although the translation of $Y_5$ produced by the ``\textit{S-NMT}'' (a context-free sentence-level NMT system) is reasonable at the sentence level, the coherence of the entire dialogue translation is poor.}
    \label{fig:case}\vspace{-8pt}
\end{figure}}

Although sentence-level Neural Machine Translation (NMT)~\cite{NIPS2014_a14ac55a,vaswani2017attention,meng2019dtmt,hassan2018achieving,yanetal2020multi,zhangetal2019bridging} has achieved promising progress, it still faces challenges in accurately translating conversational text due to abandoning the dialogue history, which leads to role-irrelevant, incoherent and inconsistent translations~\cite{mirkin-etal-2015-motivating,wang-etal-2017-semantics,laubli-etal-2018-machine,toral-etal-2018-attaining}.
Further, context-aware NMT (\citealp{tiedemann-scherrer-2017-neural,voita-etal-2018-context,voita-etal-2019-context,voita-etal-2019-good,wang-etal-2019-one,maruf-haffari-2018-document,maruf-etal-2019-selective,ma-etal-2020-simple}) can be directly applied to chat translation through incorporating the dialogue history but cannot obtain satisfactory results in this scenario~\cite{moghe-hardmeier-bawden:2020:WMT}. One important reason is the lack of explicitly modeling the inherent bilingual conversational characteristics, \emph{e.g.}, role preference, dialogue coherence, and translation consistency, as pointed out by~\citet{farajian-etal-2020-findings}. 

For a conversation, its dialogue history contains rich role preference information such as emotion, style, and humor, which is beneficial to role-relevant utterance generation~\cite{wu-etal-2020-guiding}. As shown in \autoref{fig:case}, the utterances $X_1$, $X_3$ and $X_5$ from role $R_1$ always have strong emotions (\emph{i.e.}, \emph{joy}) because of his/her preference, and preserving the same preference information across languages can help raise emotional resonance and mutual understanding~\cite{moghe-hardmeier-bawden:2020:WMT}. 
Meanwhile, there exists semantic coherence in the conversation, as the solid green arrow in \autoref{fig:case}, where the utterance $X_5$ naturally and semantically connects with the dialogue history ($X_{1\sim 4}$) on the topic ``\emph{boat}''. In addition, the bilingual conversation exhibits translation consistency, where the correct lexical choice to translate the current utterance might have appeared in preceding turns. For instance, the word ``\emph{sail}'' in $X_1$ is translated into ``\emph{ji\`achu\'{a}n}'', and thus the word ``\emph{sailing}'' in $X_3$ should be mapped into ``\emph{ji\`achu\'{a}n}'' rather than other words (\emph{e.g.}, ``\emph{h\'{a}ngx\'{i}ng}''\footnote{The words ``\emph{ji\`achu\'{a}n}'' and ``\emph{h\'{a}ngx\'{i}ng}'' express similar meaning.}) to maintain translation consistency.
On the contrary, if we ignore these characteristics, translations might be role-irrelevant, incoherent, inconsistent, and detrimental to further communication like the translation produced by the ``\textit{S-NMT}'' in \autoref{fig:case}. Although the translation is acceptable at the sentence level, it is abrupt at the bilingual conversation level. 

Apparently, how to effectively exploit these bilingual conversational characteristics is one of the core issues in chat translation. And it is challenging to implicitly capture these properties by just incorporating the complex dialogue history into encoders due to lacking the relevant information guidance~\cite{farajian-etal-2020-findings}. On the other hand, the Conditional Variational Auto-Encoder (CVAE)~\cite{NIPS2015_8d55a249} has shown its superiority in learning distributions of data properties, which is often utilized to model the diversity~\cite{zhao-etal-2017-learning}, coherence~\cite{ijcai2019-727} and users' personalities~\cite{bak-oh-2019-variational}, etc. In spite of its success, adapting it to chat translation is non-trivial, especially involving multiple tailored latent variables. 

Therefore, in this paper, we propose a model, named CPCC, to \underline{c}apture role \underline{p}reference, dialogue \underline{c}oherence, and translation \underline{c}onsistency with latent variables learned by the CVAE for neural chat translation. CPCC contains three specific latent variational modules to learn the distributions of role preference, dialogue coherence, and translation consistency, respectively. Specifically, we firstly use one role-tailored latent variable, sampled from the learned distribution conditioned only on the utterances from this role, to preserve preference. Then, we utilize another latent variable, generated by the distribution conditioned on source-language dialogue history, to maintain coherence. Finally, we leverage the last latent variable, generated by the distribution conditioned on paired bilingual conversational utterances, to keep translation consistency. As a result, these tailored latent variables allow our CPCC to produce role-specific, coherent, and consistent translations, and hence make the bilingual conversation go fluently.

We conduct experiments on WMT20 Chat Translation dataset: BConTrasT (En$\Leftrightarrow$De\footnote{English$\Leftrightarrow$German:\,En$\Leftrightarrow$De. English$\Leftrightarrow$Chinese:\,En$\Leftrightarrow$Ch.})~\cite{farajian-etal-2020-findings} and a self-collected dialogue corpus: BMELD (En$\Leftrightarrow$Ch). Results demonstrate that our model achieves consistent improvements in four directions in terms of BLEU~\cite{papineni2002bleu} and TER~\cite{snover2006study}, showing its effectiveness and generalizability. Human evaluation further suggests that our model effectively alleviates the issue of role-irrelevant, incoherent and inconsistent translations compared to other methods. Our contributions are summarized as follows:
\begin{itemize}
\item To the best of our knowledge, we are the first to incorporate the role preference, dialogue coherence, and translation consistency into neural chat translation.

\item We are the first to build a bridge between the dialogue and machine translation via conditional variational auto-encoder, which effectively models three inherent characteristics in bilingual conversation for neural chat translation.

\item Our approach gains consistent and significant performance over the standard context-aware baseline and remarkably outperforms some state-of-the-art context-aware NMT models.

\item We contribute a new bilingual dialogue corpus (BMELD, En$\Leftrightarrow$Ch) with manual translations and our codes to the research community.

\end{itemize}

\section{Background}
\subsection{Sentence-Level NMT}
\label{sec:length}
Given an input sentence $X$$=$$\{x_i\}_{i=1}^M$ with $M$ tokens, the model is asked to produce its translation $Y$$=$$\{y_i\}_{i=1}^N$ with $N$ tokens. The conditional distribution of the NMT is:
\begin{equation}\nonumber
\setlength{\abovedisplayskip}{1pt}
\setlength{\belowdisplayskip}{1pt}
\label{eq:nmt}
    p_{\theta}(Y|X) = \prod_{t=1}^{N}p_{\theta}(y_t|X, y_{1:t-1}),
\end{equation}
where $\theta$ are model parameters and $y_{1:t-1}$ is the partial translation. 

\subsection{Context-Aware NMT}
Given a source context $D_X$$=$$\{X_i\}_{i=1}^J$ and a target context $D_Y$$=$$\{Y_i\}_{i=1}^J$ with $J$ aligned sentence pairs ($X_i$, $Y_i$), the context-aware NMT~\cite{ma-etal-2020-simple} is formalized as:
\begin{equation}
\setlength{\abovedisplayskip}{1pt}
\setlength{\belowdisplayskip}{1pt}
\label{eq:dnmt}
    p_{\theta}(D_Y|D_X) = \prod_{i=1}^{J}p_{\theta}(Y_i|X_i, X_{<i}, Y_{<i}),\nonumber
\end{equation}
where $X_{<i}$ and $Y_{<i}$ are the preceding context. 
\subsection{Variational NMT}
The variational NMT model~\cite{zhang-etal-2016-variational} is the combination of CVAE~\cite{NIPS2015_8d55a249} and NMT. It introduces a random latent variable $\mathbf{z}$ into the NMT conditional distribution:
\begin{equation}
\setlength{\abovedisplayskip}{3pt}
\setlength{\belowdisplayskip}{3pt}
\label{eq:vnmt}
    p_{\theta}(Y|X) = \int_\mathbf{z} p_{\theta}(Y|X, \mathbf{z}) \cdot p_{\theta}(\mathbf{z}|X) d\mathbf{z}.
\end{equation}
Given a source sentence $X$, a latent variable $\mathbf{z}$ is firstly sampled by the prior network from the encoder, and then target sentence is generated by the decoder: $Y \sim p_{\theta}(Y|X, \mathbf{z})$, where $\mathbf{z} \sim p_{\theta}(\mathbf{z}|X)$. 

As it is hard to marginalize \autoref{eq:vnmt}, the CVAE training objective is a variational lower bound of the conditional log-likelihood:
\begin{equation}
\setlength{\abovedisplayskip}{2pt}
\setlength{\belowdisplayskip}{2pt}
\begin{split}   
\label{eq:elbo}
    \mathcal{L}(\theta,\phi;X,Y) &= -\mathrm{KL}(q_\phi (\mathbf{z}|X,Y) \| p_\theta (\mathbf{z}|X)) \nonumber \\
                       &+ \mathbb{E}_{q_\phi (\mathbf{z}|X,Y)} [\log p_\theta(Y|\mathbf{z}, X)] \\
                       & \leq \log p (Y|X), \nonumber
\end{split}
\end{equation}
where $\phi$ are parameters of the posterior network and $\mathrm{KL}(\cdot)$ indicates Kullback–Leibler divergence between two distributions produced by prior networks and posterior networks~\cite{NIPS2015_8d55a249,kingma2013auto}.

\section{Chat NMT}
We aim to learn a model that can capture inherent characteristics in the bilingual dialogue history for producing high-quality translations, \emph{i.e.}, using the context for better translations~\cite{farajian-etal-2020-findings}. Following~\cite{maruf-etal-2018-contextual}, we define paired bilingual utterances ($X_i, Y_i$) as a turn in \autoref{fig:ctx_case}, where we will translate the current utterance $X_{2k+1}$ at the $(2k+1)$-{th} turn. Here, we denote the \underline{u}tterance $X_{2k+1}$ as $X_u$ and its translation $Y_{2k+1}$ as $Y_u$ for simplicity, where $X_u$$=$$\{x_i\}_{i=1}^{m}$ with $m$ tokens and $Y_u$$=$$\{y_i\}_{i=1}^n$ with $n$ tokens. Formally, the conditional distribution for the current utterance is
\begin{equation}
\setlength{\abovedisplayskip}{1pt}
\setlength{\belowdisplayskip}{1pt}
\begin{split} 
\label{eq:cnmt}
    &p_{\theta}(Y_u|X_u, C)= \prod_{t=1}^{n}p_{\theta}(y_{t}|X_u, y_{1:t-1}, C),\nonumber
\end{split}
\end{equation}
where $C$ is the bilingual dialogue history. 

Before we dig into the details of how to utilize $C$, we define three types of context in $C$ (as shown in \autoref{fig:ctx_case}): (1) the set of previous role-specific source-language turns, denoted as $C^{role}_X$$=$$\{X_1,X_3,X_5,...,X_{2k+1}\}$\footnote{$C^{role}_Y$$=$$\{Y_2,Y_4,Y_6,...,Y_{2k}\}$ is also role-specific utterances of the interlocutor, which is used to model the interlocutor's consistency in the reverse translation direction. Here, we take one translation direction (\emph{i.e.}, En$\Rightarrow$Ch) as an example.} where $k\in[0, \frac{|T|-3}{2}]$ and $T$ is the total number of turns; (2) the set of previous source-language turns, denoted as $C_X$$=$$\{X_1,X_2,X_3,...,X_{2k}\}$; and (3) the set of previous target-language turns, denoted as $C_Y$$=$$\{Y_1,Y_2,Y_3,...,Y_{2k}\}$.
\textbf{\begin{figure}[t]
    \centering
    \includegraphics[width=0.48\textwidth]{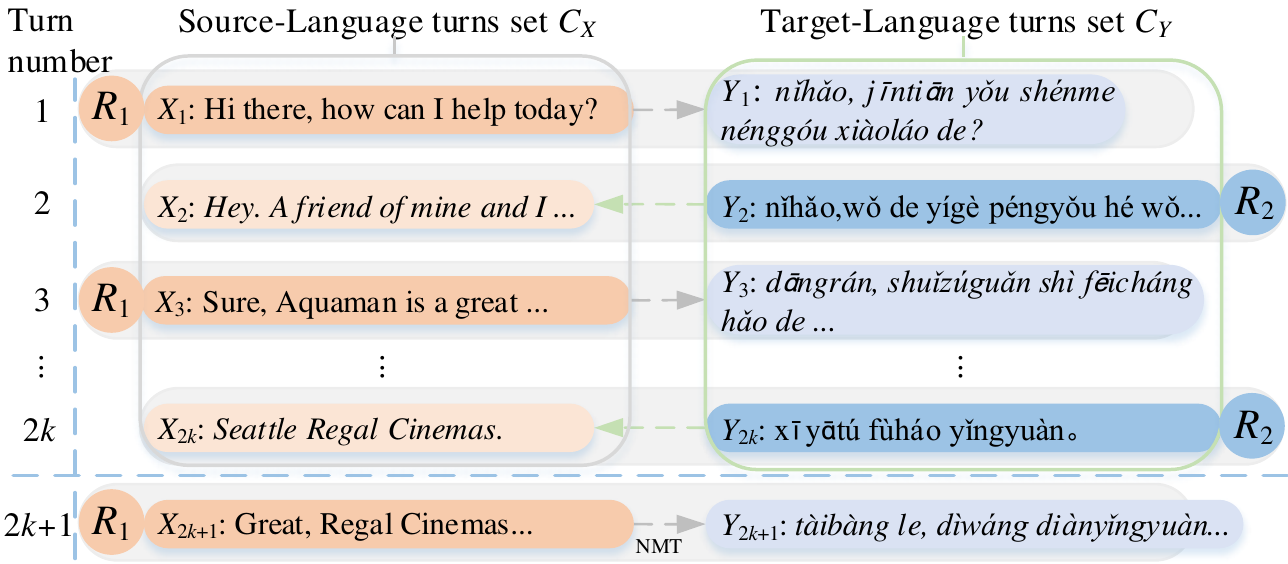}
    \caption{A dialogue example (En$\Leftrightarrow$Ch) when translating the utterance $X_{2k+1}$ where $k\in[0, \frac{|T|-1}{2}]$ and $T$ is the total number of turns (assumed to be odd here). 
    }
    \label{fig:ctx_case}\vspace{-10pt}
\end{figure}}

\begin{figure*}[ht]
\centering
  \includegraphics[width = 0.9\textwidth]{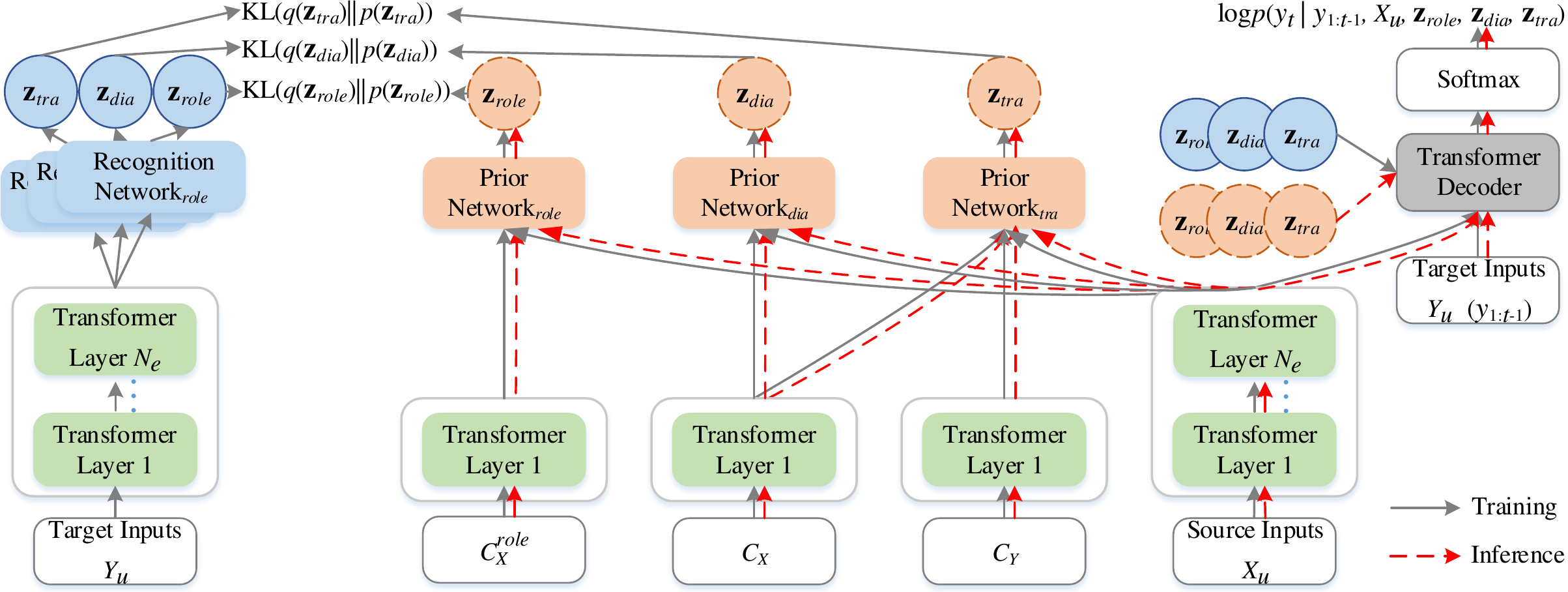}
\caption{Overview of our CPCC. The latent variables $\mathbf{z}_{role}$, $\mathbf{z}_{dia}$, and $\mathbf{z}_{tra}$ are tailored for maintaining the role preference, dialogue coherence, and translation consistency, respectively. The solid grey lines indicate training process responsible for generating \{$\mathbf{z}_{role},\mathbf{z}_{dia},\mathbf{z}_{tra}$\} from the corresponding posterior distribution predicted by recognition networks. The dashed red lines indicate inference process for generating \{$\mathbf{z}_{role},\mathbf{z}_{dia},\mathbf{z}_{tra}$\} from the corresponding prior distributions predicted by prior networks. The first Transformer layer is shared with all inputs. } 
\label{fig:Architecture}\vspace{-15pt}
\end{figure*}
\section{Our Methodology}
\autoref{fig:Architecture} demonstrates an overview of our model, consisting of five components: \emph{input representation}, \emph{encoder}, \emph{latent variational modules}, \emph{decoder}, and \emph{training objectives}. Specifically, we aim to model both dialogue and translation simultaneously. Therefore, for the \emph{input representation} (\autoref{sec:input_rep}), we incorporate dialogue-level embeddings, \emph{i.e.}, role and dialogue turn embeddings, into the \emph{encoder} (\autoref{sec:enc}). Then, we introduce three specific \emph{latent variational modules} (\autoref{sec:rsc}) to learn the distributions for varied inherent bilingual characteristics. Finally, we elaborate on how to incorporate the three tailored latent variables sampled from the distributions into the \emph{decoder} (\autoref{sec:dec}) and our two-stage \emph{training objectives} (\autoref{sec:to}).
\subsection{Input Representation}
\label{sec:input_rep}
The CPCC contains three types of inputs: source input $X_u$, target input $Y_u$, and context inputs \{$C_X^{role}$, $C_X$, $C_Y$\}. Apart from the conventional word embeddings $\mathbf{WE}$ and position embeddings $\mathbf{PE}$~\cite{vaswani2017attention}, we also introduce role embeddings $\mathbf{RE}$ and dialogue turn embeddings $\mathbf{TE}$ to identify different utterances. Specifically, for $X_u$, we firstly project it into these embeddings. Then, we perform a sum operation to unify them into a single input for each token $x_i$:
\begin{equation}
\label{emb}
\setlength{\abovedisplayskip}{2pt}
\setlength{\belowdisplayskip}{2pt}
\resizebox{.89\hsize}{!}{$
\begin{split}
    \mathbf{h}^0_i &= \mathbf{WE}({x_i}) + \mathbf{PE}({x_i}) + \mathbf{RE}({x_i}) + \mathbf{TE}({x_i}), 
\end{split}
$}
\end{equation}
where $1\leq i \leq m$ and $\mathbf{WE} \in \mathbb{R}^{|V| \times d}$, $\mathbf{RE}\in \mathbb{R}^{|R| \times d}$ and $\mathbf{SE}\in \mathbb{R}^{|T| \times d}$. $|V|$, $|R|$, $|T|$, and $d$ denote the size of shared vocabulary, number of roles, max turns of dialogue, and hidden size, respectively. $\mathbf{h}^0 \in \mathbb{R}^{m \times d}$, similarly for $Y_u$. For each of \{$C_X^{role}$, $C_X$, $C_Y$\}, we add `[cls]' tag at the head of it and use `[sep]' tag to separate its utterances~\cite{bert}, and then get its embeddings via \autoref{emb}. 

\subsection{Encoder}
\label{sec:enc}
The Transformer encoder consists of $N_e$ stacked layers and each layer includes two sub-layers:\footnote{We omit the layer normalization for simplicity, and you may refer to~\cite{vaswani2017attention} for more details.} a multi-head self-attention ($\mathrm{SelfAtt}$) sub-layer and a position-wise feed-forward network ($\mathrm{FFN}$) sub-layer~\cite{vaswani2017attention}:
\begin{equation}
\setlength{\abovedisplayskip}{2pt}
\setlength{\belowdisplayskip}{2pt}
\begin{split}
    \mathbf{s}^\ell_e &= \mathrm{SelfAtt}(\mathbf{h}^{\ell-1}_e) + \mathbf{h}^{\ell-1}_e,\ \mathbf{h}^{\ell-1}_e \in \mathbb{R}^{m \times d}, \nonumber\\
    \mathbf{h}^\ell_e &= \mathrm{FFN}(\mathbf{s}^\ell_e) + \mathbf{s}^\ell_e,\ \{\mathbf{h}^{\ell}_e, \mathbf{s}^{\ell}_e\} \in \mathbb{R}^{m \times d}, \nonumber
\end{split}
\end{equation}
where $\mathbf{h}^{\ell}_e$ denotes the state of the $\ell$-th encoder layer and $\mathbf{h}^{0}_e$ denotes the initialized feature $\mathbf{h}^{0}$. 

We prepare the representations of $X_u$ and \{$C_X^{role}$, $C_X$, $C_Y$\} for training prior and recognition networks. For $X_u$, we apply \emph{mean-pooling} with mask operation over the output $\mathbf{h}^{N_e,X}_{e}$ of the $N_e$-th encoder layer, \emph{i.e.}, $\mathbf{h}_{X}$$=$$\frac{1}{m}\sum_{i=1}^{m}(\mathbf{M}^X_i\mathbf{h}^{N_e,X}_{e,i})$, $\mathbf{h}_{X} \in \mathbb{R}^{d}$, where $\mathbf{M}^X \in \mathbb{R}^{m}$ denotes the mask matrix, whose value is either 1 or 0 indicating whether the token is padded~\cite{zhang-etal-2016-variational}. For $C_X^{role}$, as shown in \autoref{fig:Architecture}, we follow~\cite{ma-etal-2020-simple} and share the first encoder layer to obtain the context representation. Here, we take the hidden state of `[cls]' as its representation, denoted as $\mathbf{h}^{ctx}_{role}\in \mathbb{R}^{d}$. Similarly, we obtain representations of $C_X$ and $C_Y$, denoted as $\mathbf{h}^{ctx}_{X}\in \mathbb{R}^{d}$ and $\mathbf{h}^{ctx}_{Y}\in \mathbb{R}^{d}$, respectively.

For training recognition networks, we obtain the representation of $Y_u$ as $\mathbf{h}_{Y}$$=$$\frac{1}{n}\sum_{i=1}^{n}(\mathbf{M}^Y_i\mathbf{h}^{N_e,Y}_{e,i})$, $\mathbf{h}_{Y} \in \mathbb{R}^{d}$, where $\mathbf{M}^Y \in \mathbb{R}^{n}$, similar to $\mathbf{M}^X$. 

\subsection{Latent Variational Modules}
We design three tailored latent variational modules to learn the distributions of inherent bilingual conversational characteristics, \emph{i.e.}, role preference, dialogue coherence, and translation consistency. 
\label{sec:rsc}
\paragraph{Role Preference.} To preserve the role preference when translating the role's current utterance, we only encode the previous utterances of this role and produce a role-tailored latent variable $\mathbf{z}_{role}\in \mathbb{R}^{d_z}$, where ${d_z}$ is the latent size. Inspired by~\cite{ijcai2019-727}, we use isotropic Gaussian distribution as the prior distribution of $\mathbf{z}_{role}$: $p_\theta(\mathbf{z}_{role}|X_u,C^{role}_X) \sim \mathcal{N}(\bm{\mu}_{role}, \bm{\sigma}_{role}^2\mathbf{I})$, where $\mathbf{I}$ denotes the identity matrix and we have
\begin{equation}
\setlength{\abovedisplayskip}{2pt}
\setlength{\belowdisplayskip}{2pt}
\begin{split}
        \bm{\mu}_{role}&= \mathrm{MLP}_\theta^{role}(\mathbf{h}_X; \mathbf{h}^{ctx}_{role}), \nonumber \\ 
        \bm{\sigma}_{role} &=\mathrm{Softplus}(\mathrm{MLP}_\theta^{role}(\mathbf{h}_X; \mathbf{h}^{ctx}_{role})), \nonumber
\end{split}
\end{equation}
where $\mathrm{MLP(\cdot)}$ and $\mathrm{Softplus(\cdot)}$ are multi-layer perceptron and approximation of
$\mathrm{ReLU}$ function, respectively. ($\cdot$;$\cdot$) indicates concatenation operation. 

At training, the posterior distribution conditions on both role-specific utterances and the current translation, which contain rich role preference information. Therefore, the prior network can learn a role-tailored distribution by approaching the posterior network via $\mathrm{KL}$ divergence~\cite{NIPS2015_8d55a249}: $q_\phi(\mathbf{z}_{role}|X_u,C^{role}_X,Y_u)\sim \mathcal{N}({\bm \mu}_{role}^\prime, {\bm \sigma}_{role}^{{\prime}2}\mathbf{I})$ and \{${\bm \mu}_{role}^\prime$, $\bm{\sigma}_{role}^{\prime}$\} are calculated as:
\begin{equation}
\setlength{\abovedisplayskip}{5pt}
\setlength{\belowdisplayskip}{-5pt}
\begin{split}
        \bm{\mu}^\prime_{role}&= \mathrm{MLP}_\phi^{role}(\mathbf{h}_X; \mathbf{h}^{ctx}_{role}; \mathbf{h}_{Y}), \nonumber \\ 
        \bm{\sigma}^\prime_{role} &=\mathrm{Softplus}(\mathrm{MLP}_\phi^{role}(\mathbf{h}_{X}; \mathbf{h}^{ctx}_{role}; \mathbf{h}_{Y})). \nonumber
\end{split}
\end{equation}
\label{sec:lwc}
\paragraph{Dialogue Coherence.}
To maintain the coherence in chat translation, we encode the entire source-language utterances and then generate a latent variable $\mathbf{z}_{dia}\in \mathbb{R}^{d_z}$. Similar to $\mathbf{z}_{role}$, we define its prior distribution as: $p_\theta(\mathbf{z}_{dia}|X_u,C_X)\sim \mathcal{N}(\bm{\mu}_{dia}, \bm{\sigma}_{dia}^2\mathbf{I})$ and \{$\bm{\mu}_{dia}$, $\bm{\sigma}_{dia}$\} are calculated as:
\begin{equation}
\setlength{\abovedisplayskip}{-5pt}
\setlength{\belowdisplayskip}{5pt}
\begin{split}
        \bm{\mu}_{dia}&= \mathrm{MLP}_\theta^{dia}(\mathbf{h}_{X}; \mathbf{h}^{ctx}_{X}), \nonumber \\ 
        \bm{\sigma}_{dia} &=\mathrm{Softplus}(\mathrm{MLP}_\theta^{dia}(\mathbf{h}_{X}; \mathbf{h}^{ctx}_{X})). \nonumber 
\end{split}
\end{equation}

At training, the posterior distribution conditions on both the entire source-language utterances and the translation that provide a dialogue-level coherence clue, and is responsible for guiding the learning of the prior distribution. Specifically, we define the posterior distribution as: $q_\phi(\mathbf{z}_{dia}|X_u,C_X,Y_u)\sim \mathcal{N}(\bm{\mu}_{dia}^\prime, \bm{\sigma}_{dia}^{\prime2}\mathbf{I})$, where $\bm{\mu}_{dia}^\prime$ and $\bm{\sigma}_{dia}^{\prime}$ are calculated as:
\begin{equation}
\setlength{\abovedisplayskip}{5pt}
\begin{split}
        \bm{\mu}^\prime_{dia}&= \mathrm{MLP}_\phi^{dia}(\mathbf{h}_{X}; \mathbf{h}^{ctx}_{X}; \mathbf{h}_{Y}), \nonumber \\ 
        \bm{\sigma}^\prime_{dia} &=\mathrm{Softplus}(\mathrm{MLP}_\phi^{dia}(\mathbf{h}_{X}; \mathbf{h}^{ctx}_{X}; \mathbf{h}_{Y})). \nonumber
\end{split}
\end{equation}

\paragraph{Translation Consistency.}
To keep the lexical choice of translation consistent with those of previous utterances, we encode the paired source-target utterances and then sample a latent variable $\mathbf{z}_{tra}\in \mathbb{R}^{d_z}$. We define its prior distribution as: $p_\theta(\mathbf{z}_{tra}|X_u,C_X,C_Y)\sim \mathcal{N}(\bm{\mu}_{tra}, \bm{\sigma}_{tra}^2\mathbf{I})$ and \{$\bm{\mu}_{tra}$, $\bm{\sigma}_{tra}$\} are calculated as:
\begin{equation}
\setlength{\abovedisplayskip}{5pt}
\setlength{\belowdisplayskip}{5pt}
\begin{split}
        \bm{\mu}_{tra}&= \mathrm{MLP}_\theta^{tra}(\mathbf{h}_{X}; \mathbf{h}^{ctx}_{X}; \mathbf{h}^{ctx}_{Y}), \nonumber \\
        \bm{\sigma}_{tra} &=\mathrm{Softplus}(\mathrm{MLP}_\theta^{tra}(\mathbf{h}_{X}; \mathbf{h}^{ctx}_{X}; \mathbf{h}^{ctx}_{Y})). \nonumber 
\end{split}
\end{equation}

At training, the posterior distribution conditions on all paired bilingual dialogue utterances that contain implicit and aligned information, and serves as learning of the prior distribution. Specifically, we define the posterior distribution as: $q_\phi(\mathbf{z}_{tra}|X_u,C_X,C_Y,Y_u)\sim \mathcal{N}(\bm{\mu}_{tra}^\prime, \bm{\sigma}_{tra}^{\prime2}\mathbf{I})$, where $\bm{\mu}_{tra}^\prime$ and $\bm{\sigma}_{tra}^{\prime}$ are calculated as:
\begin{equation}
\setlength{\abovedisplayskip}{5pt}
\setlength{\belowdisplayskip}{5pt}
\resizebox{.99\hsize}{!}{$
\begin{split}
        \bm{\mu}^\prime_{tra}&= \mathrm{MLP}_\phi^{tra}(\mathbf{h}_{X}; \mathbf{h}^{ctx}_{X}; \mathbf{h}^{ctx}_{Y}; \mathbf{h}_{Y}), \nonumber \\ 
        \bm{\sigma}^\prime_{tra} &=\mathrm{Softplus}(\mathrm{MLP}_\phi^{tra}(\mathbf{h}_{X}; \mathbf{h}^{ctx}_{X}; \mathbf{h}^{ctx}_{Y}; \mathbf{h}_{Y})). \nonumber
\end{split}$}
\end{equation}
\subsection{Decoder}
\label{sec:dec}
The decoder adopts a similar structure to the encoder, and each of $N_d$ decoder layers contains an additional cross-attention sub-layer ($\mathrm{CrossAtt}$):
\begin{equation}
\setlength{\abovedisplayskip}{4pt}
\setlength{\belowdisplayskip}{4pt}
\begin{split}
\label{eq:trans_de}
    \mathbf{s}^\ell_d &= \mathrm{SelfAtt}(\mathbf{h}^{\ell-1}_d) + \mathbf{h}^{\ell-1}_d,\ \mathbf{h}^{\ell-1}_d \in \mathbb{R}^{n \times d},\nonumber\\
    \mathbf{c}^\ell_d &= \mathrm{CrossAtt}(\mathbf{s}^{\ell}_d, \mathbf{h}_e^{N_e}) + \mathbf{s}^\ell_d,\ \mathbf{s}^{\ell}_d \in \mathbb{R}^{n \times d},\nonumber\\
    \mathbf{h}^\ell_d &= \mathrm{FFN}(\mathbf{c}^\ell_d) + \mathbf{c}^\ell_d,\ \{\mathbf{c}^{\ell}_d, \mathbf{h}^{\ell}_d\} \in \mathbb{R}^{n \times d},\nonumber
\end{split}
\end{equation}
where $\mathbf{h}^{\ell}_d$ denotes the state of the $\ell$-th decoder layer.

As shown in \autoref{fig:Architecture}, we obtain the latent variables \{$\mathbf{z}_{role},\mathbf{z}_{dia},\mathbf{z}_{tra}$\} either from the posterior distribution predicted by recognition networks (training process as the solid grey lines) or from prior distribution predicted by prior networks (inference process as the dashed red lines). Finally, we incorporate \{$\mathbf{z}_{role},\mathbf{z}_{dia},\mathbf{z}_{tra}$\} into the state of the top layer of the decoder with a projection layer:
\begin{equation}
\setlength{\abovedisplayskip}{4pt}
\setlength{\belowdisplayskip}{4pt}
\resizebox{1.04\hsize}{!}{$
\begin{split}
    \mathbf{o}_t &= \mathrm{Tanh}(\mathbf{W}_p[\mathbf{h}^{N_d}_{d,t}; \mathbf{z}_{role};\mathbf{z}_{dia};\mathbf{z}_{tra}] + \mathbf{b}_p),\ \mathbf{o}_t \in \mathbb{R}^{d}, \nonumber
\end{split}
$}
\end{equation}
where $\mathbf{W}_p \in \mathbb{R}^{d\times (d+3d_z)}$ and $\mathbf{b}_p \in \mathbb{R}^{d}$ are training parameters, $\mathbf{h}^{N_d}_{d,t}$ is the hidden state at time-step $t$ of the $N_d$-th decoder layer. Then, $\mathbf{o}_t$ is fed to a linear transformation and softmax layer to predict the probability distribution of the next target token:
\begin{equation}
\setlength{\abovedisplayskip}{-11pt}
\setlength{\belowdisplayskip}{4pt}
\begin{split}
    \mathbf{p}_t &= \mathrm{Softmax}(\mathbf{W}_o\mathbf{o}_t+\mathbf{b}_o),\ \mathbf{p}_t \in \mathbb{R}^{|V|},\nonumber
\end{split}
\end{equation}
where $\mathbf{W}_o \in \mathbb{R}^{|V|\times d}$ and $\mathbf{b}_o \in \mathbb{R}^{|V|}$ are training parameters.

\subsection{Training Objectives}
\label{sec:to}
We apply a two-stage training strategy~\cite{zhang-etal-2018-improving,ma-etal-2020-simple}. Firstly, we train our model on large-scale sentence-level NMT data to minimize the cross-entropy objective: 
\begin{equation}\nonumber
\setlength{\abovedisplayskip}{2pt}
\setlength{\belowdisplayskip}{2pt}
\resizebox{.76\hsize}{!}{$
\begin{split}
    &\mathcal{L}(\theta;X,Y) =-\sum^{N}_{t=1}\mathrm{log}p_{\theta}(y_t|X,y_{1:t-1}).
\end{split}
$}
\end{equation}
Secondly, we fine-tune it on the chat translation data to maximize the following objective:
\begin{equation}\nonumber
\setlength{\abovedisplayskip}{4pt}
\setlength{\belowdisplayskip}{4pt}
\resizebox{.99\hsize}{!}{$
\begin{split}
    &\mathcal{J}(\theta,\phi;X_u,C_X^{role},C_X,C_Y,Y_u) =\\
    &-\mathrm{KL}(q_\phi (\mathbf{z}_{role}|X_u,C_X^{role},Y_u) \| p_\theta (\mathbf{z}_{role}|X_u,C_X^{role})) \\
    &-\mathrm{KL}(q_\phi (\mathbf{z}_{dia}|X_u,C_X,Y_u) \| p_\theta (\mathbf{z}_{dia}|X_u,C_X)) \\
    &-\mathrm{KL}(q_\phi (\mathbf{z}_{tra}|X_u,C_X,C_Y,Y_u) \| p_\theta (\mathbf{z}_{tra}|X_u,C_X,C_Y)) \\
    &+\mathbb{E}_{q_\phi} [\mathrm{log} p_\theta(Y_u|X_u, \mathbf{z}_{role}, \mathbf{z}_{dia}, \mathbf{z}_{tra})].
\end{split}
$}
\end{equation}
We use the reparameterization trick~\cite{kingma2013auto} to estimate the gradients of the prior and recognition networks~\cite{zhao-etal-2017-learning}.
\section{Experiments}
\subsection{Datasets and Metrics}
\label{sect:pdf}
\paragraph{Datasets.} We apply a two-stage training strategy, \emph{i.e.}, firstly training on a large-scale sentence-level NMT corpus (WMT20\footnote{http://www.statmt.org/wmt20/translation-task.html}) and then fine-tuning on chat translation corpus (BConTrasT~\cite{farajian-etal-2020-findings}\footnote{http://www.statmt.org/wmt20/chat-task.html} and BMELD). The details (WMT20 data and results of the first stage) are shown in Appendix A.

\paragraph{BConTrasT.} The dataset\footnote{https://github.com/Unbabel/BConTrasT} is first provided by WMT 2020 Chat Translation Task~\cite{farajian-etal-2020-findings}, which is translated from English into German and is based on the monolingual Taskmaster-1 corpus~\cite{byrne-etal-2019-taskmaster}. The conversations (originally in English) were first automatically translated into German and then manually post-edited by Unbabel editors,\footnote{www.unbabel.com} who are native German speakers. Having the conversations in both languages allows us to simulate bilingual conversations in which one speaker, the customer, speaks in German and the other speaker, the agent, answers in English.

\paragraph{BMELD.} Similarly, based on the dialogue dataset in the MELD (originally in English)~\cite{poria-etal-2019-meld},\footnote{The MELD is a multimodal emotionLines dialogue dataset, each utterance of which corresponds to a video, voice, and text, and is annotated with detailed emotion and sentiment.} we firstly crawled the corresponding Chinese translations from this\footnote{https://www.zimutiantang.com/} and then manually post-edited them according to the dialogue history by native Chinese speakers, who are post-graduate students majoring in English. Finally, following~\cite{farajian-etal-2020-findings}, we assume 50\% speakers as Chinese speakers to keep data balance for Ch$\Rightarrow$En translations and build the \underline{b}ilingual MELD (BMELD). For the Chinese, we segment the sentence using Stanford CoreNLP toolkit\footnote{https://stanfordnlp.github.io/CoreNLP/index.html}.
\begin{table}[t]
\centering
\newcommand{\tabincell}[2]{\begin{tabular}{@{}#1@{}}#2\end{tabular}}
\small
\setlength{\tabcolsep}{1.8mm}{
\begin{tabular}{l|rrr|rrr}
\toprule
\multirow{2}{*}{Dataset} & \multicolumn{3}{c|}{\# Dialogues} &  \multicolumn{3}{c}{\# Utterances} \\
&Train &Valid & Test&Train &Valid & Test\\\hline
En$\Rightarrow$De    &550&78&78 &7,629 &1,040 &1,133\\
De$\Rightarrow$En    &550&78&78  &6,216 &862 &967\\
En$\Rightarrow$Ch    &1,036&108&274  &5,560&567&1,466 \\
Ch$\Rightarrow$En    &1,036&108&274   &4,427&517&1,135 \\
\bottomrule
\end{tabular}}
\caption{Statistics of chat translation data.
}\label{datasets}
\end{table}

\paragraph{Metrics.}
For fair comparison, we use the SacreBLEU\footnote{BLEU+case.mixed+numrefs.1+smooth.exp+tok.13a+\\version.1.4.13}~\cite{post-2018-call} and v0.7.25 for TER~\cite{snover2006study} (the lower the better) with the statistical significance test~\cite{koehn-2004-statistical}. For En$\Leftrightarrow$De, we report case-sensitive score following the WMT20 chat task~\cite{farajian-etal-2020-findings}. For Ch$\Rightarrow$En, we report case-insensitive score. For En$\Rightarrow$Ch, we report the character-level BLEU score. 

\subsection{Implementation Details}
For all experiments, we follow the \emph{Transformer-Base} and \emph{Transformer-Big} settings illustrated in~\cite{vaswani2017attention}. In \emph{Transformer-Base}, we use 512 as hidden size (\emph{i.e.}, $d$), 2048 as filter size and 8 heads in multi-head attention. In \emph{Transformer-Big}, we use 1024 as hidden size, 4096 as filter size, and 16 heads in multi-head attention. All our Transformer models contain $N_e$ = 6 encoder layers and $N_d$ = 6 decoder layers and all models are trained using THUMT~\cite{tan-etal-2020-thumt} framework. We conduct experiments on the validation set of En$\Rightarrow$De to select the hyperparameters of context length and latent dimension, which are then shared for all tasks. For the results and more details (other hyperparameters setting and average running time), please refer to Appendix B, C, and D. 

\subsection{Comparison Models}
\label{ssec:layout}
\begin{table*}[t!]
\centering
\newcommand{\tabincell}[2]{\begin{tabular}{@{}#1@{}}#2\end{tabular}}
\setlength{\tabcolsep}{0.4mm}{
\begin{tabular}{c|l|cc|cc|cc|cc}
\toprule
&\multicolumn{1}{c|}{\multirow{2}{*}{\textbf{Models}}} &\multicolumn{2}{c|}{$\textbf{En$\Rightarrow$De}$}  &  \multicolumn{2}{c|}{$\textbf{De$\Rightarrow$En}$}    &\multicolumn{2}{c|}{$\textbf{En$\Rightarrow$Ch}$}  &  \multicolumn{2}{c}{$\textbf{Ch$\Rightarrow$En}$} \\ 
&\multicolumn{1}{c|}{} & \multicolumn{1}{c}{BLEU$\uparrow$} & \multicolumn{1}{c|}{TER$\downarrow$} & \multicolumn{1}{c}{BLEU$\uparrow$} &  \multicolumn{1}{c|}{TER$\downarrow$}      & \multicolumn{1}{c}{BLEU$\uparrow$} & \multicolumn{1}{c|}{TER$\downarrow$} & \multicolumn{1}{c}{BLEU$\uparrow$} & \multicolumn{1}{c}{TER$\downarrow$}   \\ \hline\hline
\multirow{2}{*}{\tabincell{c}{\emph{Baseline}\\\emph{NMT models (Base)}}}
&{Transformer}    & 40.02     & 42.5   & 48.38     & 33.4      &21.40     &72.4 & 18.52    & 59.1  \\
&{Transformer+FT}   & 58.43 & 26.7   & \underline{59.57}& 26.2 &25.22 &62.8 & 21.59 &56.7\\\cdashline{1-10}[4pt/2pt]
\multirow{3}{*}{\tabincell{c}{\emph{Context-Aware}\\\emph{NMT models (Base)}}}
&{Doc-Transformer+FT}  & 58.15     & 27.1   &59.46    &\underline{25.7}  & 24.76     &63.4  & 20.61   & 59.8   \\ 
&{Dia-Transformer+FT}  &58.33     &26.8   &59.09    &26.2   &24.96     &63.7 & 20.49   & 60.1   \\
&{V-Transformer+FT}   & \underline{58.74}   & \underline{26.3}   & 58.67   & 27.0    & \underline{26.82}     &\underline{60.6} & \underline{21.86}  & \underline{56.3}\\\cdashline{1-10}[4pt/2pt]
\multirow{1}{*}{\tabincell{c}{\emph{Ours (Base)}}}
&{CPCC}   & \textbf{60.13}$^{\dagger\dagger}$  & \textbf{25.4}$^{\dagger\dagger}$  &\textbf{61.05}$^{\dagger\dagger}$   &\textbf{24.9}$^{\dagger\dagger}$  &\textbf{27.55}$^{\dagger}$     &\textbf{60.1}$^{\dagger}$  & \textbf{22.50}$^{\dagger}$   & \textbf{55.7}$^{\dagger}$   \\\hline\hline
\multirow{2}{*}{\tabincell{c}{\emph{Baseline}\\\emph{NMT models (Big)}}}
&{Transformer}    &40.53     &42.2   &49.90 &33.3   &22.81  &69.6     &19.58    &57.7 \\
&{Transformer+FT}   &\underline{59.01} &\underline{26.0} & 59.98  &25.9   &26.95 &60.7 &22.15 &56.1\\\cdashline{1-10}[4pt/2pt]
\multirow{3}{*}{\tabincell{c}{\emph{Context-Aware}\\\emph{NMT models (Big)}}}
&{Doc-Transformer+FT}  &58.61    &26.5 &59.98 &\underline{25.4}   &26.45  &62.6   &21.38    &57.7  \\ 
&{Dia-Transformer+FT}  &58.68    &26.8  &59.63 &26.0   &26.72  &62.4    &21.09    &58.1  \\ 
&{V-Transformer+FT}   & 58.70  &26.2 &\underline{60.01}   &25.7  &\underline{27.52} &\underline{60.3}    &\underline{22.24}     &\underline{55.9}\\ \cdashline{1-10}[4pt/2pt]
\multirow{1}{*}{\tabincell{c}{\emph{Ours (Big)}}}
&{CPCC}   &\textbf{60.23}$^{\dagger\dagger}$  &\textbf{25.6}$^{\dagger}$ &\textbf{61.45}$^{\dagger\dagger}$  &\textbf{24.8}$^{\dagger}$ &\textbf{28.98}$^{\dagger\dagger}$  &\textbf{59.0}$^{\dagger\dagger}$  &\textbf{22.98}$^{\dagger}$  &\textbf{54.6}$^{\dagger\dagger}$\\ 
\bottomrule
\end{tabular}}
\caption{Results on BConTrasT (En$\Leftrightarrow$De) and BMELD (En$\Leftrightarrow$Ch) in terms of BLEU (\%) and TER (\%). The best and the second results are bold and underlined, respectively. ``$^{\dagger}$'' and ``$^{\dagger\dagger}$'' indicate that statistically significant better than the best result of all contrast NMT models with t-test {\em p} \textless \ 0.05 and {\em p} \textless \ 0.01, respectively.}
\label{tbl:main_res}\vspace{-5pt}
\end{table*}
\paragraph{Baseline NMT Models.}
{Transformer}~\cite{vaswani2017attention}: the de-facto NMT model that does not fine-tune on chat translation data. {Transformer+FT}: fine-tuning on the chat translation data after being pre-trained on sentence-level NMT corpus. 

\paragraph{Context-Aware NMT Models.}
{Doc-Transformer+FT}~\cite{ma-etal-2020-simple}: a state-of-the-art document-level NMT model based on Transformer sharing the first encoder layer to incorporate the bilingual dialogue history. {Dia-Transformer+FT}~\cite{maruf-etal-2018-contextual}: using an additional RNN-based~\cite{hochreiter1997long} encoder to incorporate the mixed-language dialogue history, where we re-implement it based on Transformer and use another Transformer layer to introduce context. {V-Transformer+FT}~\cite{zhang-etal-2016-variational,mccarthy-etal-2020-addressing}: the variational NMT model based on Transformer also sharing the first encoder layer to exploit the bilingual context for fair comparison.

\begin{table}[t!]
\centering
\scalebox{0.84}{
\setlength{\tabcolsep}{0.30mm}{
\begin{tabular}{l|l|cc|cc}
\toprule
\multirow{2}{*}{\#}&\multicolumn{1}{c|}{\multirow{2}{*}{\textbf{Models}}} &\multicolumn{2}{c|}{$\textbf{En$\Rightarrow$De}$}  &  \multicolumn{2}{c}{$\textbf{De$\Rightarrow$En}$}    \\ 
&\multicolumn{1}{c|}{} & \multicolumn{1}{c}{BLEU$\uparrow$} & \multicolumn{1}{c|}{TER$\downarrow$} & \multicolumn{1}{c}{BLEU$\uparrow$} & \multicolumn{1}{c}{TER$\downarrow$}   \\ \hline
0&{CPCC (\emph{Base})}   & \textbf{60.96}  & \textbf{24.6} &\textbf{62.09}    & \textbf{24.5} \\\hline 
1&\emph{w}/\emph{o} {$\mathbf{z}_{role}$ } & 60.56 (-0.40)   & 25.1    & 61.42 (-0.67) & 24.8      \\ 
2&\emph{w}/\emph{o} {$\mathbf{z}_{dia}$}  & 60.50 (-0.46)  & 25.2    &61.65 (-0.44)   & 25.1  \\ 
3&\emph{w}/\emph{o} {$\mathbf{z}_{tra}$}    & 60.39 (-0.57)  & 25.1    &61.38 (-0.71)    &26.0 \\
4&\emph{w}/\emph{o}  $\mathbf{z}_{role}$ \& $\mathbf{z}_{dia}$ & 59.64 (-1.32)  & 25.8  & 60.65 (-1.44) & 25.8 \\
5&\emph{w}/\emph{o} $\mathbf{z}_{role}$ \& $\mathbf{z}_{tra}$  & 59.61 (-1.35)  & 25.9  & 60.62 (-1.47) & 25.7 \\
6&\emph{w}/\emph{o} $\mathbf{z}_{dia}$ \& $\mathbf{z}_{tra}$  & 60.24 (-0.72)  & 25.1   &61.18 (-0.91)    & 24.9 \\
7&\emph{w}/\emph{o} {all}   & 58.95 (-2.01)     & 26.1     & 59.82 (-2.27)    & 26.1   \\
\bottomrule
\end{tabular}}}
\caption{Ablation study on the validation set. ``\emph{w/o} all'' indicates removing all latent variables but remaining encoding all bilingual dialogue history.}
\label{tbl:ablation}
\end{table}
\subsection{Main Results}
Overall, we separate the models into two parts in \autoref{tbl:main_res}: the \emph{Base} setting and the \emph{Big} setting. In each part, we show the results of our re-implemented Transformer baselines, the context-aware NMT systems, and our approach on En$\Leftrightarrow$De and En$\Leftrightarrow$Ch. 
\paragraph{Results on En$\Leftrightarrow$De.}
\label{ssec:ende}
Under the \emph{Base} setting, CPCC substantially outperforms the baselines (\emph{e.g.}, ``Transformer+FT'') by a large margin with 1.70$\uparrow$ and 1.48$\uparrow$ BLEU scores on En$\Rightarrow$De and De$\Rightarrow$En, respectively. On the TER, our CPCC achieves a significant improvement of 1.3 points in both language pairs. Under the \emph{Big} setting, our CPCC also consistently boosts the performance in both directions (\emph{i.e.}, 1.22$\uparrow$ and 1.47$\uparrow$ BLEU scores, 0.4$\downarrow$ and 1.1$\downarrow$ TER scores), showing its effectiveness.

Compared against the strong context-aware NMT systems (underlined results), our CPCC significantly surpasses them (about 1.39$\sim$1.59$\uparrow$ BLEU scores and 0.6$\sim$0.9$\downarrow$ TER scores) in both language directions under both \emph{Base} and \emph{Big} settings, demonstrating the superiority of our model.

\paragraph{Results on En$\Leftrightarrow$Ch.}
\label{ssec:chen}
We also conduct experiments on our self-collected data to validate the generalizability across languages in \autoref{tbl:main_res}.

Our CPCC presents remarkable BLEU improvements over the ``Transformer+FT'' by a large margin in two directions by 2.33$\uparrow$ and 0.91$\uparrow$ BLEU gains under the \emph{Base} setting, respectively, and by 2.03$\uparrow$ and 0.83$\uparrow$ BLEU gains in both directions under the \emph{Big} setting. These results suggest that CPCC consistently performs well across languages.

Compared with strong context-aware NMT systems (\emph{e.g.}, ``V-Transformer+FT''), our approach notably surpasses them in both language directions under both \emph{Base} and \emph{Big} settings, which shows the generalizability and superiority of our model.

\section{Analysis}
\subsection{Ablation Study}
\label{ssec:abs}
We conduct ablation studies to investigate how well each tailored latent variable of our model works. When removing latent variables listed in \autoref{tbl:ablation}, we have the following findings. 

(1) All latent variables make substantial contributions to performance, proving the importance of modeling role preference, dialogue coherence, and translation consistency, which is consistent with our intuition that the properties should be beneficial to better translations (rows 1$\sim$3 vs. row 0).

(2) Results of rows 4$\sim$7 show the combination effect of three latent variables, suggesting that the combination among three latent variables has a cumulative effect (rows 4$\sim$7 vs. rows 0$\sim$3). 

(3) Row 7 vs. row 0 shows that explicitly modeling the bilingual conversational characteristics significantly outperforms implicit modeling (\emph{i.e.}, just incorporating the dialogue history into encoders), which lacks the relevant information guidance.  

\subsection{Dialogue Coherence}
Following~\cite{lapata2005automatic,Xiong_He_Wu_Wang_2019}, we measure dialogue coherence as sentence similarity. Specifically, the representation of each sentence is the mean of the distributed vectors of its words, and the dialogue coherence between two sentences $s_1$ and $s_2$ is determined by the cosine similarity:
\begin{equation}\nonumber
\setlength{\abovedisplayskip}{5pt}
\setlength{\belowdisplayskip}{5pt}
\begin{split}  
    sim(s_1, s_2) &= cos(f({s_1}), f({s_2})),\\ 
    f(s_i) &= \frac{1}{\vert s_i\vert}\sum_{\mathbf{w} \in s_i}(\mathbf{w}),
\end{split}
\end{equation}
where \(\mathbf{w}\) is the vector for word \(w\).

We use Word2Vec\footnote{https://code.google.com/archive/p/word2vec/}~\cite{mikolov2013efficient} to learn the distributed vectors of words by training on the monolingual dialogue dataset: Taskmaster-1~\cite{byrne-etal-2019-taskmaster}. And we set the dimensionality of word embeddings to 100.

\autoref{coherence} shows the cosine similarity on the test set of De$\Rightarrow$En. It reveals that our model encouraged by tailor-made latent variables produces better coherence in chat translation than contrast systems.

\label{ssec:dc}
\begin{table}[t]
\centering
\newcommand{\tabincell}[2]{\begin{tabular}{@{}#1@{}}#2\end{tabular}}
\setlength{\tabcolsep}{0.4mm}{
\begin{tabular}{l|c|c|c}
\toprule
\multirow{1}{*}{\textbf{Models}} &  \multicolumn{1}{c|}{\textbf{1-th Pr.}} & \multicolumn{1}{c|}{\textbf{2-th Pr.}}  & \multicolumn{1}{c}{\textbf{3-th Pr.}}\\\cline{1-4}
Transformer               &0.6502  &0.6037 &0.5659\\
Transformer+FT            &0.6587  &0.6104 &0.5714  \\\hline
Doc-Transformer+FT        &0.6569  &0.6093 &0.5713\\
Dia-Transformer+FT        &0.6553  &0.6084 &0.5709\\
V-Transformer+FT          &0.6602  &0.6122 &0.5751\\\hline
CPCC (Ours)  &0.6660$^{\dagger\dagger}$  &\textbf{0.6190}$^{\dagger\dagger}$ &\textbf{0.5814}$^{\dagger\dagger}$  \\
Human Reference         &\textbf{0.6663}  &\textbf{0.6190} &0.5795\\
\bottomrule
\end{tabular}}
\caption{Results of dialogue coherence in terms of sentence similarity ({De$\Rightarrow$En}, \emph{Base}). The ``\#\textbf{-th Pr.}'' denotes the \#-th preceding utterance to the current one. ``$^{\dagger\dagger}$'' indicates that statistically significant better than the best result of all contrast NMT models ({\em p} \textless \ 0.01).}\label{coherence} 
\end{table}


\subsection{Human Evaluation}
\label{ssec:he}
Inspired by~\cite{bao-EtAl:2020:WMT,farajian-etal-2020-findings}, we use four criteria for human evaluation: (1) \textbf{Preference} measures whether the translation preserves the role preference information; (2) \textbf{Coherence} denotes whether the translation is semantically coherent with the dialogue history; (3) \textbf{Consistency} measures whether the lexical choice of translation is consistent with the preceding utterances; (4) \textbf{Fluency} measures whether the translation is logically reasonable and grammatically correct.

We firstly randomly sample 200 examples from the test set of Ch$\Rightarrow$En. Then, we assign each bilingual dialogue history and corresponding 6 generated translations to three human annotators without order, and ask them to evaluate whether each translation meets the criteria defined above. All annotators are postgraduate students and not involved in other parts of our experiments.

\autoref{human_evaluation} shows that our CPCC effectively alleviates the problem of role-irrelevant, incoherent and inconsistent translations compared with other models (significance test~\cite{koehn-2004-statistical}, {\em p} \textless \ 0.05), indicating the superiority of our model. The inter-annotator agreement is 0.527, 0.491, 0.556 and 0.485 calculated by the Fleiss’ kappa~\cite{doi:10.1177/001316447303300309}, for {preference}, {coherence}, {consistency} and {fluency}, respectively, indicating ``Moderate Agreement'' for all four criteria. We also present some case studies in Appendix H.
\begin{table}[t]
\centering
\newcommand{\tabincell}[2]{\begin{tabular}{@{}#1@{}}#2\end{tabular}}
\setlength{\tabcolsep}{0.8mm}{
\begin{tabular}{l|c|c|c|c}
\toprule
\multirow{1}{*}{\textbf{Models}} & \multicolumn{1}{c|}{$\textbf{Pref.}$} &  \multicolumn{1}{c|}{$\textbf{Coh.}$} & \multicolumn{1}{c|}{$\textbf{Con.}$}& \multicolumn{1}{c}{$\textbf{Flu.}$} \\\cline{1-5}
Transformer        &0.485  &0.540 &0.510  &0.590 \\
Transformer+FT     &0.530  &0.590 &0.565  &0.635 \\\hline
Doc-Transformer+FT &0.525  &0.595 &0.560  &0.630 \\
Dia-Transformer+FT &0.525  &0.580 &0.555  &0.625 \\
V-Transformer+FT   &0.535  &0.595 &0.560  &0.635 \\\hline 
CPCC (Ours)        &\textbf{0.570}  &\textbf{0.620} &\textbf{0.585}  &\textbf{0.650} \\
\bottomrule
\end{tabular}}
\caption{Results of Human evaluation ({Ch$\Rightarrow$En}, \emph{Base}). ``\textbf{Pref.}'': Preference. ``\textbf{Coh.}'': Coherence. ``\textbf{Con.}'': Consistency. ``\textbf{Flu.}'': Fluency. }\label{human_evaluation} 
\end{table}
\label{ssec:cs}
\section{Related Work}
\paragraph{Chat NMT.} It only involves several researches due to the lack of human-annotated publicly available data~\cite{farajian-etal-2020-findings}. Therefore, some existing work~\cite{lrec,maruf-etal-2018-contextual,9023129,rikters-etal-2020-document} mainly pays attention to designing methods to automatically construct the subtitles corpus, which may contain noisy bilingual utterances. Recently, ~\citet{farajian-etal-2020-findings} organize the WMT20 chat translation task and first provide a human post-edited corpus, where some teams investigate the effect of dialogue history and finally ensemble their models for higher ranks (\citealp{berard-EtAl:2020:WMT,mohammed-alayyoub-abdullah:2020:WMT,wang-EtAl:2020:WMT1,bao-EtAl:2020:WMT,moghe-hardmeier-bawden:2020:WMT}). As a synchronizing study,~\citet{wang2021autocorrect} use multitask learning to auto-correct the translation error, such as pronoun dropping, punctuation dropping, and typos. Unlike them, we focus on explicitly modeling role preference, dialogue coherence, and translation consistency with tailored latent variables to promote the translation quality. 

\paragraph{Context-Aware NMT.}\quad Chat NMT can be viewed as a special case of context-aware NMT, which has attracted many researchers \cite{gong-etal-2011-cache,jean2017does,wang-etal-2017-exploiting-cross,bawden-etal-2018-evaluating,miculicich-etal-2018-document,articleKuang,tu-etal-2018-learning,yang-etal-2019-enhancing,kang-etal-2020-dynamic,li-etal-2020-multi,ma-etal-2020-simple} to extend the encoder or decoder for exploring the context impact on translation quality. Although these models can be directly applied to chat translation, they cannot explicitly capture the bilingual conversational characteristics and thus lead to unsatisfactory translations~\cite{moghe-hardmeier-bawden:2020:WMT}. Different from these studies, we focus on explicitly modeling these bilingual conversational characteristics via CVAE for better translations. 

\paragraph{Conditional Variational Auto-Encoder.}\quad CVAE has verified its superiority in many fields~\cite{NIPS2015_8d55a249}. In NMT, \citet{zhang-etal-2016-variational} and \citet{Su_Wu_Xiong_Lu_Han_Zhang_2018} extend CVAE to capture the global/local information of source sentence for better results. \citet{mccarthy-etal-2020-addressing} focus on addressing the posterior collapse with mutual information. Besides, some studies use CVAE to model the correlations between image and text for multimodal NMT~\cite{toyama2016neural,calixto-etal-2019-latent}. Although the CVAE has been widely used in NLP tasks, its adaption and utilization to chat translation for modeling inherent bilingual conversational characteristics are non-trivial, and to the best of our knowledge, has never been investigated before.

\section{Conclusion and Future Work}
We propose to model bilingual conversational characteristics through tailored latent variables for neural chat translation. Experiments on En$\Leftrightarrow$De and En$\Leftrightarrow$Ch directions show that our model notably improves translation quality on both BLEU and TER metrics, showing its superiority and generalizability. Human evaluation further verifies that our model yields role-specific, coherent, and consistent translations by incorporating tailored latent variables into NMT. Moreover, we contribute a new bilingual dialogue data (BMELD, En$\Leftrightarrow$Ch) with manual translations to the research community. In the future, we would like to explore the effect of multimodality and emotion on chat translation, which has been well studied in dialogue field~\cite{liang2020infusing}.

\section*{Acknowledgments}
The research work descried in this paper has been supported by the National Key R\&D Program of China (2020AAA0108001) and the National Nature Science Foundation of China (No. 61976015, 61976016, 61876198 and  61370130). The authors would like to thank the anonymous reviewers for their valuable comments and suggestions to improve this paper.

\bibliographystyle{acl_natbib}
\bibliography{acl2021}
\appendix
\label{sec:appendix}
\section*{Appendix}
\section{Datasets}
\paragraph{WMT20.} For the En$\Leftrightarrow$De, we combine six corpora including Euporal, ParaCrawl, CommonCrawl, TildeRapid, NewsCommentary, and WikiMatrix, and we combine News Commentary v15, Wiki Titles v2, UN Parallel Corpus V1.0, CCMT Corpus, and WikiMatrix for the En$\Leftrightarrow$Ch. We firstly filter noisy sentence pairs according to their characteristics in terms of duplication and length  (whose length exceeds 80). To pre-process the raw data, we employ a series of open-source/in-house scripts, including full-/half-width conversion, unicode conversation, punctuation normalization, and tokenization~\cite{wang-EtAl:2020:WMT1}. After filtering steps, we generate subwords via joint BPE~\cite{sennrich-etal-2016-neural} with 32K merge operations. Finally, we obtain 45,541,367 sentence pairs for En$\Leftrightarrow$De and 22,244,006 sentence pairs for En$\Leftrightarrow$Ch, respectively. 

We test the model performance of the first stage on \emph{newstest2019}. The results are shown in \autoref{bleu_on_one_stage}.


\begin{table}[t]
\centering
\newcommand{\tabincell}[2]{\begin{tabular}{@{}#1@{}}#2\end{tabular}}
\small
\setlength{\tabcolsep}{0.5mm}{
\begin{tabular}{l|l|c|c|c|c}
\toprule
&\multirow{1}{*}{Methods} & \multicolumn{1}{c|}{$\textbf{En$\Rightarrow$De}$} &  \multicolumn{1}{c|}{$\textbf{De$\Rightarrow$En}$} & \multicolumn{1}{c|}{$\textbf{En$\Rightarrow$Ch}$} &  \multicolumn{1}{c}{$\textbf{Ch$\Rightarrow$En}$} \\\cline{1-6}

\multirow{2}{*}{\tabincell{c}{\emph{Base}}}
&Transformer  &39.88  &40.72  &32.55  &24.42\\
&V-Transformer   &40.01  &41.36  &32.90 &25.77\\\hline
\multirow{2}{*}{\tabincell{c}{\emph{Big}}}
&Transformer   &41.35  &41.56 &33.85  &24.86\\
&V-Transformer   &41.40  &41.67  &33.90  &26.46\\
\bottomrule
\end{tabular}}
\caption{The BLEU scores on the \emph{newstest2019} of the first stage. 
}\label{bleu_on_one_stage} 
\end{table}

\section{Implementation Details}
For all experiments, we follow two model settings illustrated in~\cite{vaswani2017attention}, namely \emph{Transformer-Base} and \emph{Transformer-Big}. The training step is set to 200,000 and 2,000 for the first stage and the fine-tuning stage, respectively. The batch size for each GPU is set to 4096 tokens. The beam size is set to 4, and the length penalty is 0.6 among all experiments. All experiments in the first stage are conducted utilizing 8 NVIDIA Tesla V100 GPUs, while we use 2 GPUs for the second stage, \emph{i.e.}, fine-tuning. That gives us about 8*4096 and 2*4096 tokens per update for all experiments in the first-stage and second-stage, respectively. All models are optimized using Adam~\cite{kingma2017adam} with $\beta_1$ = 0.9 and $\beta_2$ = 0.998, and learning rate is set to 1.0 for all experiments. Label smoothing is set to 0.1. We use dropout of 0.1/0.3 for \emph{Base} and \emph{Big} setting, respectively. To alleviate the degeneration problem of the variational framework, we apply KL annealing. The KL multiplier $\lambda $ gradually increases from 0 to 1 over 10, 000 steps. $|R|$ is set to 2 for En$\Leftrightarrow$De and 7 for En$\Leftrightarrow$Ch, respectively. $|T|$ is set to 10.
The criterion for selecting hyperparameters is the BLEU score on validation sets for both tasks. The average running time is shown in \autoref{time_on_one_stage}. 

In the case of blind testing or online use (assumed dealing with En$\Rightarrow$De), since translations of target utterances (\emph{i.e.}, English) will not be given, an inverse De$\Rightarrow$En model is simultaneously trained and used to back-translate target utterances~\cite{bao-EtAl:2020:WMT}, similar to all tasks.

\begin{table}[t]
\centering
\newcommand{\tabincell}[2]{\begin{tabular}{@{}#1@{}}#2\end{tabular}}
\small
\setlength{\tabcolsep}{0.3mm}{
\begin{tabular}{l|l|c|c|c|c}
\toprule
&\multirow{1}{*}{Stages} & \multicolumn{1}{c|}{$\textbf{En$\Rightarrow$De}$} &  \multicolumn{1}{c|}{$\textbf{De$\Rightarrow$En}$} & \multicolumn{1}{c|}{$\textbf{En$\Rightarrow$Ch}$} &  \multicolumn{1}{c}{$\textbf{Ch$\Rightarrow$En}$} \\\cline{1-6}

\multirow{2}{*}{\tabincell{c}{\emph{Base}}}
&The First Stage  &5D  &7D  &4D  &3.5D\\
&Fine-Tuning Stage   &4H  &5H  &3H &2H\\\hline
\multirow{2}{*}{\tabincell{c}{\emph{Big}}}
&The First Stage   &10D  &12D &7D  &6D\\
&Fine-Tuning Stage  &4.5H  &5.5H  &4H  &2.5H\\
\bottomrule
\end{tabular}}
\caption{The average running time for the first stage and fine-tuning stage. D: Days, H: Hours.
}\label{time_on_one_stage} 
\end{table}

\begin{figure}[t]
    \centering
    \includegraphics[width=0.48\textwidth]{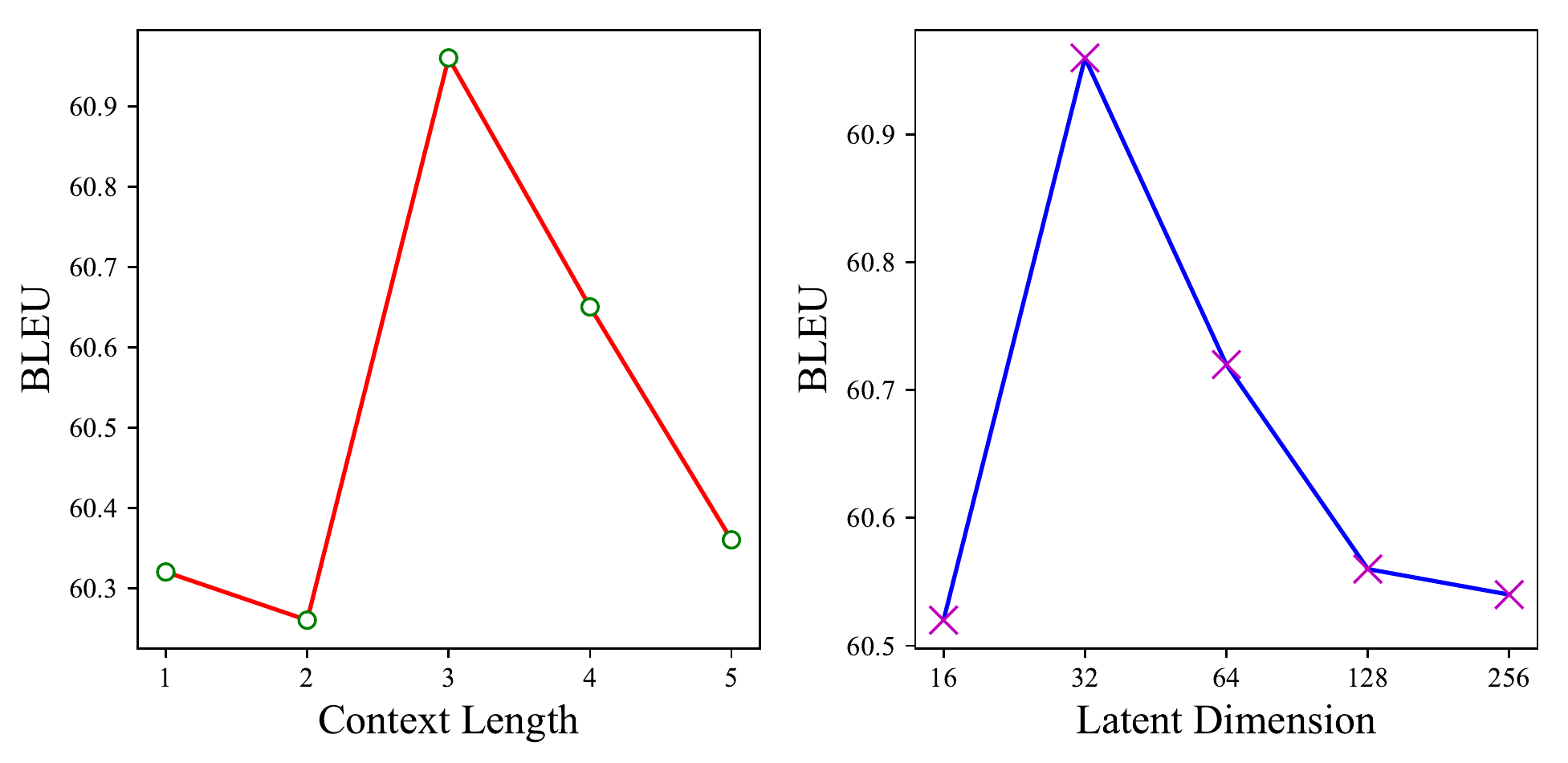}
    \caption{Effect of context length and latent dimension on translation quality. The BLEU scores (\%) are calculated on the validation set of the En$\Rightarrow$De.
    }
    \label{fig:ctx_dev_exp}
\end{figure}
\section{Effect of Context Length}
We firstly investigate the effect of context length (\emph{i.e.}, the number of preceding utterances) on our approach under the Transformer \emph{Base} setting. As shown in the left of \autoref{fig:ctx_dev_exp}, using three preceding source sentences as dialogue history achieves the best translation performance on the validation set (En$\Rightarrow$De). Using more preceding sentences does not bring any improvement and increases the computational cost. This confirms the finding of
~\citet{tu-etal-2018-learning} and ~\citet{zhang-etal-2018-improving} that long-distance context only has limited influence. Therefore, we set the number of preceding sentences to 3 in all experiments.

\section{Effect of Latent Dimension} 
The right of \autoref{fig:ctx_dev_exp} shows the effect of the latent dimension on translation quality under the Transformer \emph{Base} setting. Obviously, using latent dimension 32 suffices to achieve superior performance. Increasing the dimension does not lead to any improvements. Therefore, we set the latent dimension to 32 in all experiments.

\begin{figure}[t]
    \centering
    \includegraphics[width=0.30\textwidth]{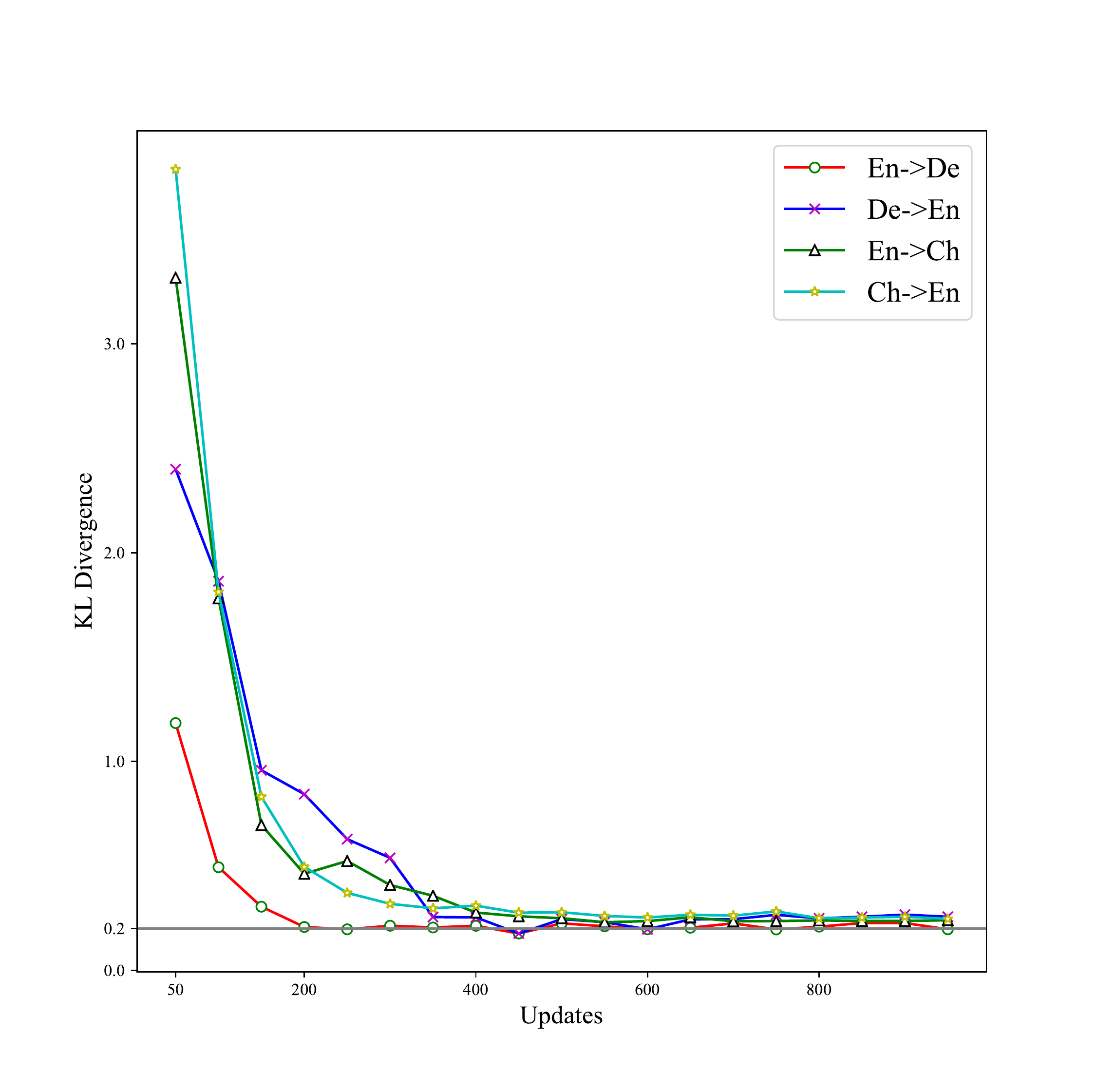}
    \caption{Total KL divergence (per word) of all latent variables (first 1,000 updates on corresponding validation set).
    }
    \label{fig:kl}
\end{figure}
\section{KL Divergence}
\label{ssec:kl}
Generally, KL divergence measures the amount of information encoded in a latent variable. In the extreme case where the KL divergence of latent variable $\mathbf{z}$ equals to zero, the model completely ignores $\mathbf{z}$, \emph{i.e.}, it degenerates. \autoref{fig:kl} shows that the total KL divergence of our model maintains around 0.2$\sim$0.5 indicating that the degeneration problem does not exist in our model and latent variables can play their corresponding roles.
\begin{figure}[t]
    \centering
    \includegraphics[width=0.48\textwidth]{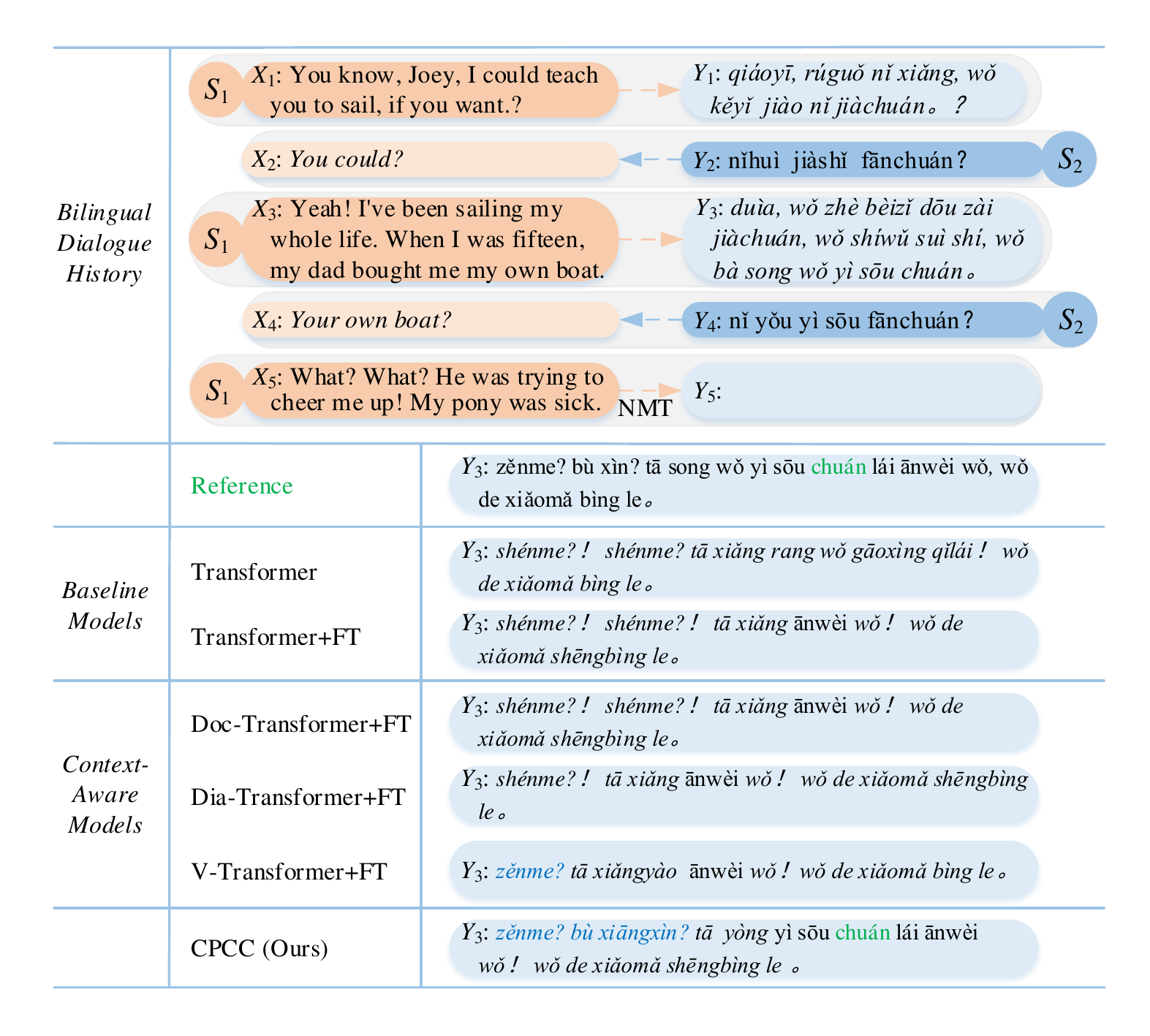}
    \caption{Bilingual conversational example one.
    }
    \label{fig:k2}
\end{figure}
\begin{figure}[t]
    \centering
    \includegraphics[width=0.48\textwidth]{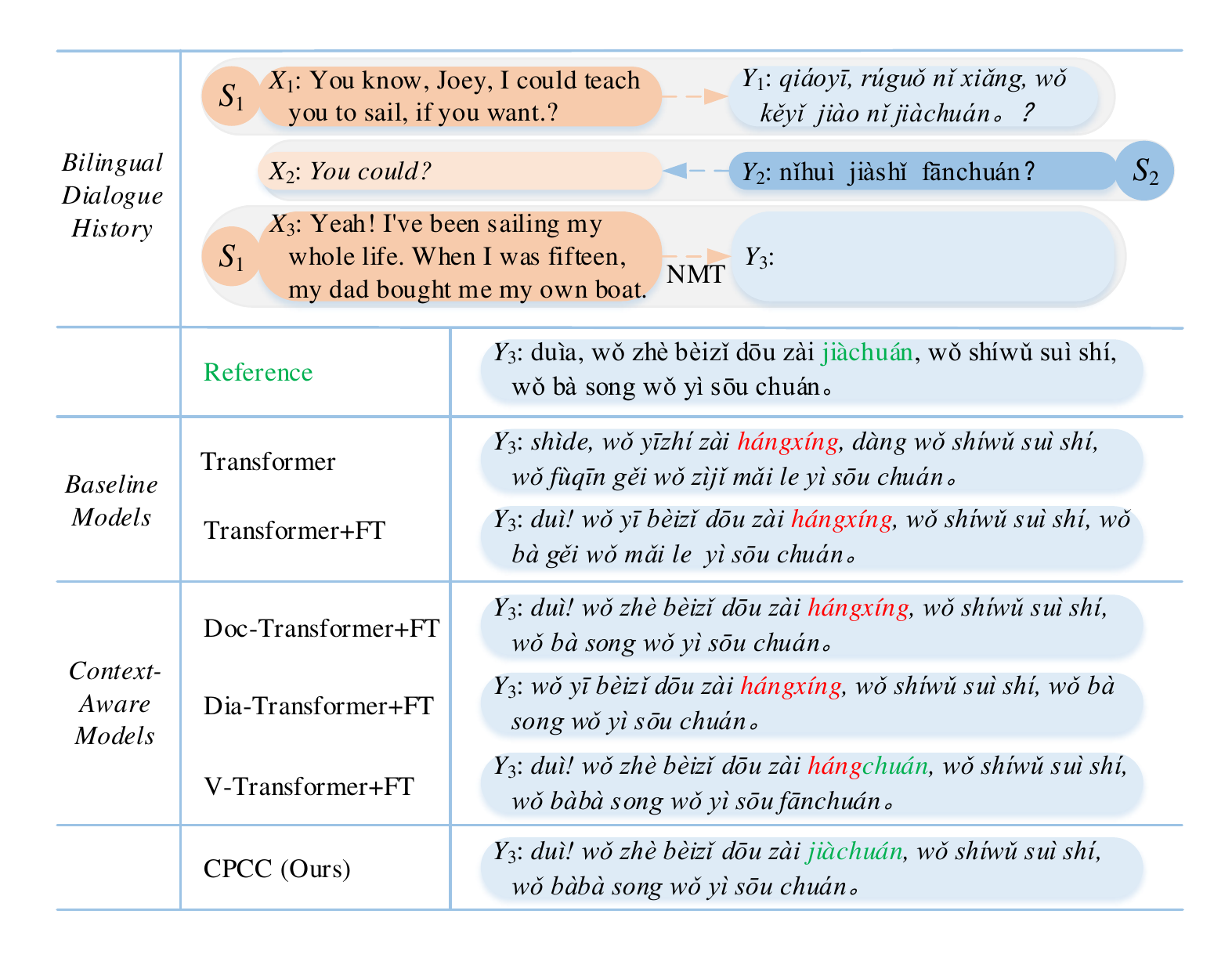}
    \caption{Bilingual conversational example two.
    }
    \label{fig:k3}
\end{figure}
\section{Case Study}
In this section, we show some cases in \autoref{fig:k2} and \autoref{fig:k3} to investigate the effect of different models. 
\paragraph{Role Preference and Dialogue Coherence.}
As shown in \autoref{fig:k2}, we observe that the baseline models and the context-aware models except ``V-Transformer+FT'' cannot preserve the role preference information, \emph{e.g.}, \emph{joy} emotion, even these ``*-Transformer+FT'' models incorporate the bilingual conversational history into the encoder. The ``V-Transformer+FT'' model produces very slightly emotional elements (\emph{e.g.}, ``\emph{z\v{e}nme?}'') due to the latent variable over the source sentence capturing relevant preference information. Meanwhile, we find that all comparison models cannot generate a coherent translation. The reason may be that they fail to capture the conversation-level coherence clue, \emph{i.e.}, ``\emph{boat}''. By contrast, we explicitly model the two characteristics through tailored latent variables and thus obtain satisfactory results.

\paragraph{Translation Consistency.}
As shown in \autoref{fig:k3}, we observe that all comparison models cannot maintain the translation consistency due to the lack of explicitly modeling this characteristic. Our model has the ability to overcome the issue and can keep the correct lexical choice to translate the current utterance that might have appeared in preceding turns, \emph{i.e.}, ``\emph{ji\`achu\`an}''.

To sum up, both cases show that our model yields role-specific, coherent, and consistent translations by incorporating tailored latent variables into translators, demonstrating its effectiveness and superiority.

\end{document}